%% file: main.tex
\documentclass{article}
\usepackage{iclr2027_conference,times}
\input{math_commands.tex}

\usepackage{amsmath,amssymb,mathtools}
\usepackage{booktabs}
\usepackage{multirow}
\usepackage{array}
\usepackage{makecell}
\usepackage{tabularx}
\usepackage{longtable}
\usepackage{graphicx}
\usepackage{wrapfig}
\usepackage{subcaption}
\usepackage{float}
\usepackage{placeins}
\usepackage{chngcntr}
\usepackage{bbm}

\usepackage{microtype}
\usepackage{inconsolata}
\usepackage{latexsym}

\usepackage[table]{xcolor}
\usepackage{url}
\usepackage{hyperref}

\usepackage{amsthm}
\theoremstyle{definition}
\newtheorem{definition}{Definition}

\usepackage{fancyhdr}

\fancypagestyle{arxivstyle}{
  \fancyhf{}
  \renewcommand{\headrulewidth}{0pt}
  \renewcommand{\footrulewidth}{0pt}
  \cfoot{\thepage}
}

\hypersetup{
  hidelinks,
  pdftitle={
    Phantom Merge: When Your Large Language Model Agents
    Pick One but Tell You About Another
  },
  pdfauthor={
    Wangshu Zhu, Qianwen Guo, Yushun Dong
  }
}

\newcommand{\makearxivtitle}{
  \begin{center}

    \rule{\textwidth}{3pt}
    \vspace{-0.1em}
    {\LARGE\bfseries
    Phantom Merge: 
    When Your Large Language Model\\ Agents Pick One
    but Tell You About Another\par}
    \vspace{-0.1em}
    \rule{\textwidth}{0.8pt}

    {\large
    Wangshu Zhu$^{\dagger}$ \quad
    Qianwen Guo$^{\ddagger}$ \quad
    Yushun Dong$^{\dagger}$\par}

    \vspace{-0.4em}

    $^{\dagger}$Department of Computer Science,
    Florida State University

    $^{\ddagger}$Department of Civil and Environmental Engineering,
    FAMU--FSU College of Engineering

    \vspace{-0.35em}

    {\ttfamily
    \{wz26b,yushun.dong\}@fsu.edu
    \quad
    qguo@eng.famu.fsu.edu\par}

  \end{center}

}

\iclrfinalcopy

\begin{document}

\pagestyle{arxivstyle}
\thispagestyle{arxivstyle}

\makearxivtitle
\raggedbottom

\input{sections/00_abstract}

\input{sections/01_introduction}

\input{sections/02_problem}

\input{sections/03_method}
\input{sections/04_setup}
\input{sections/05_experiments}
\input{sections/06_related_work}

\input{sections/07_conclusion}

% =========================================================
% Statements
% =========================================================

\clearpage

% =========================================================
% References
% =========================================================

\bibliography{references}
\bibliographystyle{iclr2027_conference}

% =========================================================
% Appendix
% =========================================================

\clearpage
\appendix
\input{sections/appendix}

\end{document}

%% file: math_commands.tex
\usepackage{amsmath,amsfonts,bm}

\def\eqref#1{equation~\ref{#1}}
\def\1{\bm{1}}

\DeclareMathAlphabet{\mathsfit}{\encodingdefault}{\sfdefault}{m}{sl}
\SetMathAlphabet{\mathsfit}{bold}{\encodingdefault}{\sfdefault}{bx}{n}

%% file: sections/00_abstract.tex
\begin{abstract}
LLM agents increasingly operate over multiple entities, where reliable response generation requires not only retrieving relevant evidence but also preserving its association with the correct entity. We identify \emph{Phantom Merge} (PM), an entity--evidence binding failure in which an agent attributes unsupported claims to its committed entity, potentially even when the task is completed correctly. We formalize PM at the claim level and characterize its prevalence and error subtypes across four agent domains and four backbone families. To address this problem, we propose a two-stage framework for claim-level risk estimation and correction. 
First, Anchor-Grounded Risk (AGR) combines a representational readout of the agent's pre-emission hidden state with Jacobian-based functional information, providing complementary slot-level and anchor-value risk scores. Second, Fixed-Anchor Correction (FAC) audits flagged claims and selectively revises them using observed anchor evidence or removes them when no supported replacement is available.
On the held-out Shopping detection cohort, AGR-slot achieves an AUROC of 96.2\% and an average precision of 97.3\%, outperforming the evaluated detection baselines. In the end-to-end Shopping mitigation evaluation, FAC reduces trajectory-level PM from 51.4\% to 6.2\%, with an output-to-original claim ratio of 80.7\%.
These results demonstrate the prevalence of entity-specific grounding failures and the effectiveness of combining internal risk estimation with evidence-constrained claim correction.
Our code is publicly available at \url{https://github.com/LabRAI/PhantomMerge}.
\end{abstract}

%% file: sections/01_introduction.tex
\section{Introduction}
\label{sec:introduction}

LLM agents extend language models from single-turn generation to interactive systems that interleave reasoning, tool calls, and observations from external environments~\citep{yao2023react,nakano2021webgpt}. This capability allows agents to search, compare, and act over multiple external entities before producing a final response~\citep{lee2025fhiragentbench}. In such settings, reliable response generation requires each final-answer claim to be supported by observations associated with the entity that the claim describes. We refer to the entity to which the agent commits in its final response as the \emph{anchor}. However, during final-answer synthesis, an agent may attribute to the anchor a value observed for another entity, project a requested constraint onto the anchor as an observed attribute, or generate a value that conflicts with the anchor's observed evidence. We define this class of entity--evidence binding failures as \emph{Phantom Merge} (PM), where an anchor-attributed claim lacks support from the anchor's own observed evidence. PM differs from generic hallucination because problematic content can already be supported elsewhere in the trajectory while remaining unsupported for the anchor. Accordingly, task-level success and trajectory-level factual support provide only partial information about whether the final response is correctly grounded in the committed entity.

However, existing approaches leave two limitations in establishing and restoring entity-specific grounding. \textit{Gap 1. Entity-specific support verification.} Current agent evaluations~\citep{liu2023agentbench} primarily assess task completion or entity selection, while factuality methods examine whether generated claims are supported by available evidence~\citep{rashkin2023attribution}. These approaches do not explicitly verify whether the supporting evidence belongs to the entity referenced by each final-answer claim. Consequently, the first challenge is to assess claim-level grounding across evidence from multiple entities collected during successive agent interactions. \textit{Gap 2. Entity-specific claim correction.} Existing post-hoc factuality methods revise or remove unsupported content~\citep{dhuliawala2024chain}, but generally do not explicitly constrain corrections to evidence associated with the entity already selected by the agent. Therefore, the second challenge is to correct incorrectly attributed claims by entity-specific evidence while preserving committed entity and correctly grounded content.

To address these challenges, we propose a two-stage framework for diagnosing and correcting Phantom Merge. For the first challenge, we introduce \emph{Anchor-Grounded Risk} (AGR), which combines two complementary readouts of the agent's pre-emission hidden state. Specifically, a representational readout estimates claim-level binding risk, while a Jacobian-based functional readout evaluates downstream compatibility with anchor-derived targets at attribute or value granularity. 
We develop a fixed-additive slot-level score for primary detection and a calibrated anchor-value configuration for complementary risk estimation. For the second challenge, we introduce \emph{Fixed-Anchor Correction}, which combines claim selection, functional auditing, and evidence-grounded revision to correct claims associated with the committed entity. On completed agent trajectories, high-risk claims are revised when a supported replacement is recoverable and removed otherwise, while unaffected claims and the committed entity remain unchanged without updating the underlying agent.

\paragraph{Contributions.}
Our contributions are summarized as follows:
\begin{itemize}
\item \textbf{A novel agent reliability problem.} We formalize Phantom Merge as an entity--evidence binding failure and systematically characterize its prevalence, subtype composition, and relationship with task correctness across agent domains and backbones.

\item \textbf{An anchor-grounded diagnosis method.}
We introduce AGR, combining representational and Jacobian-based
functional readouts for claim-level PM detection, and demonstrate
strong discrimination on Shopping, including test claims with
non-overlapping quote templates.

\item \textbf{An evidence-grounded correction method.} We develop Fixed-Anchor Correction and evaluate its effectiveness in reducing PM and preserving response content through end-to-end and shared-gate experiments across multiple agent domains.
\end{itemize}

%% file: sections/02_problem.tex
\section{Problem Definition}
\label{sec:problem}

We consider an LLM-agent trajectory $\tau=(q,x_{1:T},y)$, comprising a user query $q$, interactions and observations $x_{1:T}$, and final response $y$. Let $V_\tau$ denote the observed entities and $\mathrm{Obs}_\tau(e)$ the evidence observed for entity $e$. The agent commits to an \emph{anchor} $a_\tau\in V_\tau$, and its response contains anchor-attributed claims $C_\tau=\{c_1,\ldots,c_n\}$, each asserting a value $v(c)$ for attribute $\mathrm{slot}(c)$. We write $\mathrm{Obs}_\tau(e)\models c$ when the evidence supports $c$, and use $\mathrm{Src}_\tau(c)\subseteq V_\tau\cup\{q\}$ for the identifiable sources of its asserted value, including user constraints from $q$. 

\begin{definition}[Phantom Merge]
\label{def:phantom-merge}
A claim $c\in C_\tau$ exhibits \emph{Phantom Merge} (PM) when it lacks anchor-specific support and satisfies at least one of three conditions: \textsc{CEM} (Cross-Entity Misattribution), where a non-anchor entity supports $c$; \textsc{CAP} (Constraint-to-Anchor Projection), where a user constraint is asserted as an unsupported anchor attribute; or \textsc{ASF} (Anchor-Slot Fabrication), where neither preceding condition applies and the claim contradicts an observed anchor value. \textsc{CEM} and \textsc{CAP} may co-occur. Let $\mathcal{L}_\tau(c)$ denote the applicable labels and $\mathcal{M}=\{\textsc{CEM},\textsc{CAP},\textsc{ASF}\}$. We define
\begin{equation}
\mathrm{PM}(c)=\mathbbm{1}\!\left[\mathcal{L}_\tau(c)\cap\mathcal{M}\neq\varnothing\right].
\label{eq:pm-definition}
\end{equation}
A trajectory exhibits PM if any $c\in C_\tau$ satisfies $\mathrm{PM}(c)=1$.
\end{definition}

\begin{definition}[Pure Hallucination]
\label{def:pure-hallucination}
A claim is a \emph{Pure Hallucination} (\textsc{PH}) when $\mathrm{Src}_\tau(c)=\varnothing$ and it does not contradict any observed anchor value for $\mathrm{slot}(c)$. \textsc{PH} is excluded from PM.
\end{definition}

%% file: sections/03_method.tex
\section{Methodology}
\label{sec:method}

\begin{figure*}[t]
\centering
\includegraphics[width=\textwidth]{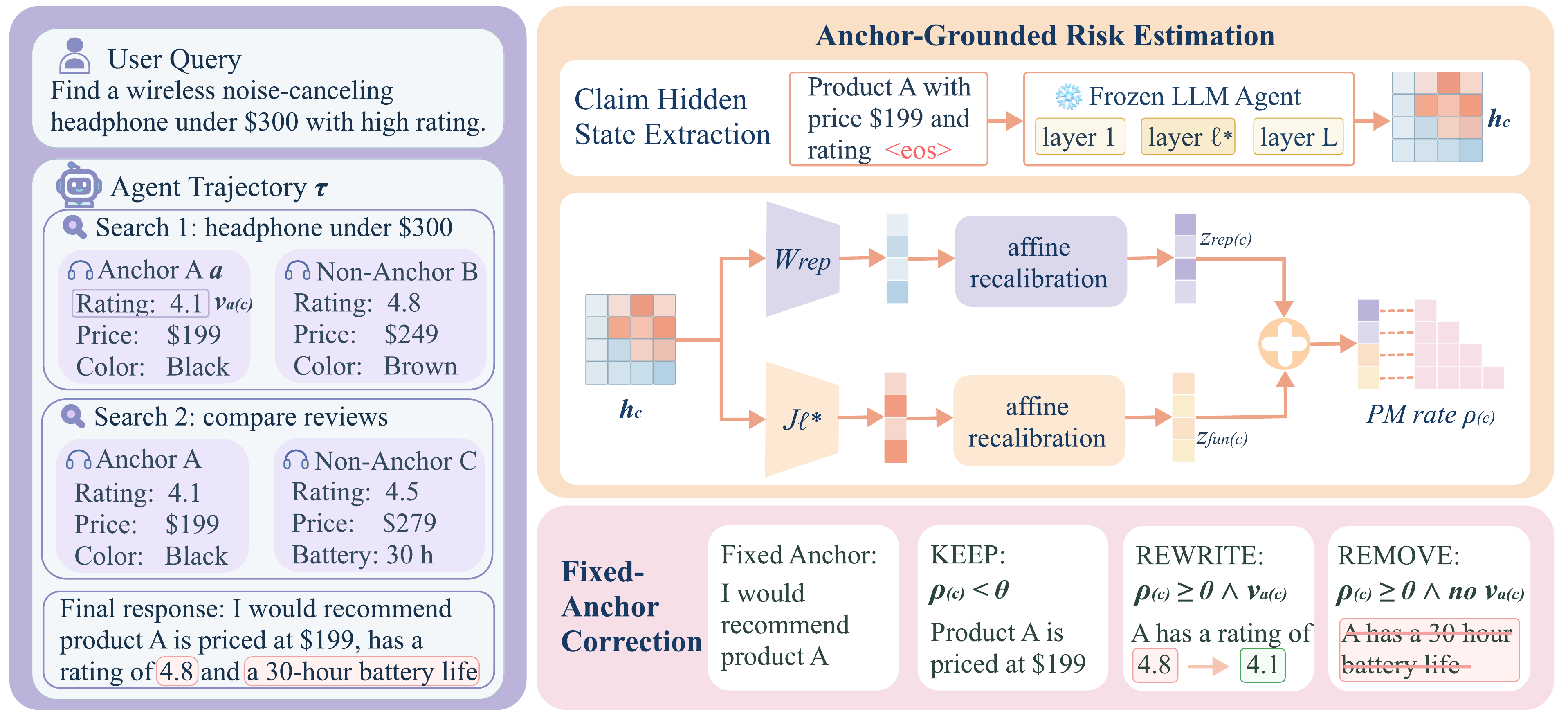}
\vspace{-6mm}
\caption{\textbf{Overview of the proposed framework.} Anchor-Grounded Risk (AGR) combines representational and Jacobian-based functional information for claim-level PM detection. Fixed-Anchor Correction (FAC) integrates risk-based selection, functional auditing, and anchor-specific evidence to retain, revise, or remove individual claims.}
\label{fig:method_overview}
\vspace{-0.65cm}
\end{figure*}

\begingroup
\setlength{\abovedisplayskip}{5pt}
\setlength{\belowdisplayskip}{5pt}
\setlength{\abovedisplayshortskip}{3pt}
\setlength{\belowdisplayshortskip}{3pt}

\subsection{Overview}
\label{sec:method-overview}

We propose a two-stage framework for diagnosing and correcting Phantom Merge, as illustrated in Figure~\ref{fig:method_overview}. Given an agent trajectory $\tau$, its committed anchor $a_\tau$, and anchor-attributed claims $C_\tau$, the first stage introduces \emph{Anchor-Grounded Risk} (AGR), which integrates representational and Jacobian-based functional information from the agent's pre-emission state. We develop two complementary scoring configurations using attribute-level and anchor-value functional targets. The second stage introduces \emph{Fixed-Anchor Correction} (FAC), which combines claim-level risk assessment with functional auditing and observed anchor evidence to determine appropriate corrections. 

\subsection{Anchor-Grounded Risk Estimation}
\label{sec:method-agr}

We estimate claim-level PM risk from the agent's internal state during final-answer generation. Let $t_c$ denote the position immediately preceding the first token of $v(c)$, and let $h_c\in\mathbb{R}^d$ be the corresponding hidden state at layer $\ell^\ast$, selected on validation data. We extract two complementary signals from this frozen representation: a supervised readout of PM-related information and a functional readout of the model's downstream computation.

\paragraph{Representational Readout.}
Given the claim-level label $m_c=\mathrm{PM}(c)\in\{0,1\}$, we train a linear readout $z_{\mathrm{rep}}(c)=w_{\mathrm{rep}}^\top h_c+b_{\mathrm{rep}}$, with $p_{\mathrm{rep}}(c)=\sigma(z_{\mathrm{rep}}(c))$, using binary cross-entropy on labeled anchor--claim instances. The restricted linear formulation measures the accessibility of PM-related information in the frozen representation~\citep{hewitt2019designing,belinkov2022probing}. The resulting representational signal provides the supervised component of AGR. We denote the output of the frozen representational probe used in the slot-level configuration by $p_L(c)$.

\paragraph{Jacobian-Based Functional Readout.}
The functional readout estimates the information accessible through the model's downstream computation. We construct two functional targets: an attribute lexicon $T_{\mathrm{slot}}(c)$ associated with $\mathrm{slot}(c)$ and an anchor-value target $T_{\mathrm{val}}(c)$ constructed from $\mathrm{Obs}_\tau(a_\tau)$. The former captures slot-level information, while the latter evaluates compatibility with the observed value of the committed anchor.

Let $F_{\ell^\ast\rightarrow L}$ denote the frozen computation from layer $\ell^\ast$ to the final residual representation at position $t_c$. We abbreviate this mapping as $F$ and approximate its output using an averaged Jacobian estimated from unlabeled generation states. With reference state $\bar h$, the resulting approximation is
\begin{equation}
\begin{aligned}
\widehat h_c^{\,L} &= F(\bar h)+J_{\ell^\ast}(h_c-\bar h),\\[-1pt]
J_{\ell^\ast} &= \mathbb{E}_{h}\!\left[\nabla_h F(h)\right].
\end{aligned}
\label{eq:jacobian-linearization}
\end{equation}
Following the Jacobian-lens approach~\citep{gurnee_verbalizable_2026}, we obtain the approximated output logits $\widehat g_c=W_U\widehat h_c^{\,L}$ using the unembedding matrix $W_U$. Let $\pi_J(T\mid h_c)$ denote the resulting probability mass assigned to token set $T$. This provides a shared functional representation from which we construct the two AGR scoring configurations.

Our primary configuration, \emph{AGR-slot}, combines the frozen representational probe with the Jacobian-based slot-level signal. Therefore, we define
\begin{equation}
B(c)=\operatorname{logit}\!\left(p_L(c)\right)+\log\pi_J(T_{\mathrm{slot}}(c)\mid h_c).
\label{eq:agr-slot}
\end{equation}
The score multiplies probe-derived PM odds by slot-level probability
mass in log space, downweighting claims with low slot compatibility.
This fixed fusion requires no additional weights and ranks claim-level PM risk.
Slot mass modulates the probe evidence rather than indicating PM on its own.

Our complementary configuration, \emph{AGR-value}, uses the anchor-value target to estimate functional compatibility with the committed entity's observed evidence. We transform the corresponding probability mass into a risk-oriented score:
\begin{equation}
z_{\mathrm{fun}}^{\mathrm{val}}(c)=
\log\frac{1-\pi_J(T_{\mathrm{val}}(c)\mid h_c)}
{\pi_J(T_{\mathrm{val}}(c)\mid h_c)}.
\label{eq:functional-readout}
\end{equation}
This formulation assigns higher functional risk to lower estimated compatibility with the observed anchor value. The value-level signal is computed for claims with recoverable anchor-value targets and subsequently integrated with the representational readout.

\paragraph{Calibrated Risk Fusion.}
AGR-value integrates the representational and anchor-value functional signals through calibrated fusion. We first apply affine calibration $\widetilde z_i=a_i z_i+b_i$ to each available signal on a held-out split $\mathcal{D}_c$, where $i\in\{\mathrm{rep},\mathrm{fun}\}$. For claims with both signals available, let $s(c)=[\widetilde z_{\mathrm{rep}}(c),\widetilde z_{\mathrm{fun}}^{\mathrm{val}}(c)]^\top$. Assuming class-conditional Gaussian scores $s\mid m_c=k\sim\mathcal N(\mu_k,\Sigma)$ with shared covariance $\Sigma$, the corresponding Bayes log-posterior odds are
\begin{equation}
\log\frac{P(m_c=1\mid s)}{P(m_c=0\mid s)}
=(\mu_1-\mu_0)^\top\Sigma^{-1}s+b.
\label{eq:agr-discriminant}
\end{equation}
This formulation motivates an affine combination of the two calibrated signals that incorporates their class separation and covariance. Under diagonal covariance, each signal's coefficient is proportional to its class-conditional mean difference divided by its variance. We define the AGR-value score as
\begin{equation}
\rho(c)=
\begin{cases}
\sigma\!\left(w^\top s(c)+b\right), & T_{\mathrm{val}}(c)\text{ resolves},\\[2pt]
\sigma\!\left(\widetilde z_{\mathrm{rep}}(c)\right), & \text{otherwise}.
\end{cases}
\label{eq:agr}
\end{equation}
The fusion parameters $(w,b)$ are fitted on a held-out split $\mathcal{D}_f$. Probe fitting, calibration, and fusion use disjoint data partitions, while the Jacobian operator is estimated independently from unlabeled states. The value-level configuration uses the representational signal when an anchor-value target is unavailable. Together, AGR-slot and AGR-value provide two implementations of representational--functional risk estimation, combining their respective targets through fixed additive scoring and calibrated fusion for complementary claim-level PM detection.

\subsection{Fixed-Anchor Claim Correction}
\label{sec:method-correction}

We introduce \emph{Fixed-Anchor Correction} (FAC) to revise risky claims using evidence associated with the committed entity. FAC consists of two components: a claim-selection gate and an internal functional audit. The gate identifies claims requiring intervention, while the audit provides additional information for selecting an appropriate correction. For each claim $c$, we retrieve the value of $\mathrm{slot}(c)$ from $\mathrm{Obs}_\tau(a_\tau)$ and denote the recovered anchor-supported value by $v_a(c)$ when available.

\paragraph{Anchor-Constrained Routing.}
Let $g(c)$ denote the score supplied by the claim-selection gate, and let $u_c$ denote the functional audit state. The audit combines representational risk, slot-level functional information, and attribute-alignment features to distinguish different intervention conditions. Given threshold $\theta$, the resulting claim-level operator is
\begin{equation}
\mathcal{A}_{\theta}(c)=
\begin{cases}
c, & g(c)\leq\theta,\\[2pt]
\operatorname{Route}\!\left(c,u_c,\mathrm{Obs}_\tau(a_\tau)\right),
& g(c)>\theta.
\end{cases}
\label{eq:correction-rule}
\end{equation}
The routing procedure uses the functional audit to assess flagged claims and determine whether to retain, revise, or remove them. Claims eligible for revision enter an evidence-grounded rewriting procedure, which uses a recovered anchor-supported value to correct the asserted attribute while preserving the remaining claim content. Claims requiring removal are excluded from the corrected response. Applying the operator to the final-answer claims produces $y'$ while preserving the committed anchor and the completed interaction trajectory.

FAC supports alternative risk estimators through its claim-selection interface. In the Shopping shared-gate comparison, all correction methods
receive the same claims selected by the frozen representational
probe, $p_{\mathrm{pm}}(c)>0.05$.
FAC then applies functional auditing and anchor-constrained
correction to these claims. We separately examine alternative AGR detection heads under fixed downstream routing to evaluate how risk estimation influences intervention outcomes.

\endgroup

%% file: sections/04_setup.tex
\section{Experiment Settings}
\label{sec:settings}

\paragraph{Benchmark.}
We evaluate Phantom Merge on four multi-entity agent domains:
ShoppingBench~\citep{wang2025shoppingbench} and the airline,
telecom, and retail domains of $\tau^3$-Bench~
\citep{yao_$t$bench_2024a,barres_$t^2$bench_2025a}.
PM characterization uses 2,000 PM-enriched Shopping trajectories
and 1,000 trajectories from each $\tau^3$-Bench domain.
Detection and mitigation additionally use an independently
annotated Shopping cohort, with a held-out test set of
243 trajectories and 274 claims (156 PM and 118 correctly bound).
The three $\tau^3$-Bench domains also serve as cross-domain
evaluation sets.

\paragraph{Baselines.}
For detection, we compare AGR with TF-IDF, two ModernBERT
classifiers~\citep{warner_smarter_2025}, SelfCheckGPT~
\citep{manakul2023selfcheckgpt}, FacTool~
\citep{chern_factool_2023}, and FActScore~
\citep{min2023factscore}, alongside individual readouts
and AGR-value.
Additional anchor-aware baselines include DeBERTa-NLI~
\citep{he_deberta_2021}, LLM verifiers, and cite-or-drop.
For mitigation, we compare Fixed-Anchor Correction with
Fixed-Anchor Deletion, RARR~\citep{gao2023rarr},
CoVe~\citep{dhuliawala2024chain}, and a FEVER-style
baseline~\citep{thorne2018fever}, under both complete
intervention and shared-gate settings.
Implementation details are provided in
Appendix~\ref{app:baseline-details}.

\paragraph{Metrics.}
For PM characterization, we report trajectory-level
prevalence, task correctness, conditional statistics,
and claim-level subtype composition.
Detection metrics include AUROC, average precision (AP),
precision, recall, F1, and recall-matched precision.
Mitigation metrics include trajectory-level PM rate,
correctly bound claim retention, and the output-to-original
claim ratio to quantify mitigation and content preservation.
Post-intervention outputs are re-evaluated using the same
AGR procedure, with manual spot checks of the resulting assessments.

\paragraph{Backbones.}
Qwen3-32B~\citep{yang2025qwen3} is our primary agent
backbone for characterization, detection, and mitigation.
We additionally evaluate Gemma-4-31B~
\citep{team_gemma_2026a}, Llama-3.3-70B~
\citep{grattafiori_llama_2024a}, and Mixtral-8x7B~
\citep{jiang_mixtral_2024a} for PM characterization.
Qwen3-8B performs deterministic claim rewriting.
All agent backbones remain frozen.

%% file: sections/05_experiments.tex
\section{Experimental Evaluations}
\label{sec:experiments}

We evaluate Phantom Merge and the proposed framework through three research
questions.
\textbf{RQ1:} How does Phantom Merge manifest across agent domains and model
backbones?
\textbf{RQ2:} How effectively can AGR detect Phantom Merge, and what contributes
to its performance?
\textbf{RQ3:}  How effectively can Fixed-Anchor Correction reduce Phantom Merge while preserving response content?

\subsection{Prevalence and Characteristics of Phantom Merge}
\label{sec:rq1}

We examine Phantom Merge across four agent domains and four model backbones, measuring its trajectory-level prevalence, occurrence among task-correct trajectories, and claim-level subtype composition. Table~\ref{tab:rq1} summarizes the results.

\begin{table*}[t]
\centering
\caption{Phantom Merge across four agent domains and backbone families.
TC denotes task correctness, PM denotes trajectory-level Phantom Merge,
and $X\mid Y$ denotes $P(X\mid Y)$. CEM, CAP, and ASF report claim-level
PM subtype composition. All values are percentages.}
\label{tab:rq1}
\vspace{-0.25cm}

\setlength{\tabcolsep}{0pt}
\renewcommand{\arraystretch}{1.08}

\begin{tabular*}{\textwidth}
{@{\extracolsep{\fill}}llccccccc@{}}
\toprule
Domain & Backbone & TC & PM & PM$\mid$TC & TC$\mid$PM
& CEM & CAP & ASF \\
\midrule

\multirow{4}{*}{Retail}
& Mixtral & 8.7 & 16.5 & 18.4 & 9.7 & 4.2 & 1.6 & 94.2 \\
& Gemma   & 21.8 & 0.4 & 0.5 & 25.0 & 0.0 & 0.0 & 100.0 \\
& Llama   & 9.2 & 2.5 & 0.0 & 0.0 & 3.6 & 7.1 & 89.3 \\
& Qwen    & 15.9 & 3.1 & 1.3 & 6.5 & 69.0 & 0.0 & 31.0 \\
\midrule

\multirow{4}{*}{Airline}
& Mixtral & 35.6 & 34.0 & 32.0 & 33.5 & 0.0 & 0.3 & 99.7 \\
& Gemma   & 29.3 & 4.5 & 4.1 & 26.7 & 0.0 & 75.0 & 25.0 \\
& Llama   & 16.8 & 5.1 & 2.4 & 7.8 & 0.0 & 7.7 & 92.3 \\
& Qwen    & 13.7 & 18.6 & 5.1 & 3.8 & 26.4 & 30.6 & 43.1 \\
\midrule

\multirow{4}{*}{Telecom}
& Mixtral & 4.0 & 11.7 & 7.7 & 2.7 & 0.0 & 0.0 & 100.0 \\
& Gemma   & 11.6 & 0.6 & 0.9 & 16.7 & 62.5 & 0.0 & 37.5 \\
& Llama   & 3.6 & 27.4 & 5.6 & 0.7 & 100.0 & 0.0 & 0.0 \\
& Qwen    & 4.1 & 54.1 & 17.1 & 1.3 & 100.0 & 0.0 & 0.0 \\
\midrule

\multirow{4}{*}{Shopping}
& Mixtral & 41.0 & 33.0 & 37.1 & 46.0 & 10.9 & 85.2 & 3.9 \\
& Gemma   & 27.7 & 41.7 & 45.5 & 30.3 & 2.6 & 95.9 & 1.5 \\
& Llama   & 35.1 & 26.8 & 29.6 & 38.8 & 1.6 & 96.4 & 2.0 \\
& Qwen    & 20.7 & 46.8 & 55.6 & 24.6 & 29.5 & 63.3 & 7.2 \\
\bottomrule
\end{tabular*}
\vspace{-0.5cm}
\end{table*}

\vspace{-0.3cm}
\paragraph{Prevalence and task correctness.}
PM occurs in all 16 evaluated backbone-domain combinations, spanning dense and mixture-of-experts models and both recommendation and customer-service tasks. Its prevalence ranges from 0.4\% for Gemma on Retail to 54.1\% for Qwen on Telecom, indicating substantial variation across configurations. More importantly, PM is not confined to unsuccessful trajectories. In the PM-enriched Shopping cohort, 29.6--55.6\% of task-correct trajectories still exhibit PM across the four backbones. This discrepancy also appears outside Shopping: PM affects 32.0\% of task-correct trajectories for Mixtral on Airline and 17.1\% for Qwen on Telecom, although it is absent among task-correct trajectories for Llama on Retail. These findings show that successful task completion can coexist with incorrectly attributed claims. Conventional task correctness therefore cannot fully capture the PM of final responses, motivating explicit claim-level assessment.

\vspace{-0.3cm}
\paragraph{Error composition.}
PM subtype composition varies markedly across both domains and backbones. Shopping is dominated by Constraint-to-Anchor Projection (63.3--96.4\%), whereas Telecom exhibits particularly pronounced backbone-dependent differences: all observed PM claims are classified as Cross-Entity Misattribution for Qwen and Llama, but as Anchor-Slot Fabrication for Mixtral. These extremes occur at substantially different PM prevalence rates, including 54.1\% for Qwen and 11.7\% for Mixtral. Airline also illustrates this heterogeneity: Anchor-Slot Fabrication accounts for 99.7\% of Mixtral's PM claims, while Qwen exhibits a more balanced distribution across all three subtypes. Importantly, extreme subtype proportions must be interpreted alongside PM prevalence and the number of observed PM claims. For instance, Gemma's 100\% Anchor-Slot Fabrication on Retail accompanies a PM prevalence of only 0.4\%, making its subtype estimate particularly sensitive to the limited number of PM instances. These results establish that PM is neither restricted to a single error subtype nor characterized by a uniform error distribution across agent configurations.

\subsection{Detection Effectiveness and Readout Analysis}
\label{sec:rq2}

We answer \textbf{RQ2} by comparing AGR with lexical, embedding-based,
and adapted factual verification methods on the same held-out Shopping
cohort. We evaluate overall detection performance and examine how the
representational and functional readouts contribute to AGR.

\begin{table*}[t]
\centering
\caption{Phantom Merge detection on the Shopping test cohort. AUROC and AUPRC are reported as percentages with 95\% bootstrap confidence intervals. Precision, recall, and F1 are evaluated at $\tau=0.05$ for applicable probability-like scores. “–” denotes scores evaluated separately using development-selected thresholds (Appendix~\ref{app:threshold-selection}).}
\label{tab:rq2}
\vspace{-0.25cm}

\setlength{\tabcolsep}{0pt}
\renewcommand{\arraystretch}{1.08}

\begin{tabular*}{\textwidth}
{@{\extracolsep{\fill}}lccccc@{}}
\toprule
Method & AUROC & AUPRC & Precision & Recall & F1 \\
\midrule

RARR
& 49.9 [48.7, 50.8]
& 56.9 [50.2, 63.0]
& 50.0 & 0.6 & 1.3 \\

FActScore
& 58.2 [51.6, 64.6]
& 65.5 [57.5, 73.0]
& 59.4 & 62.8 & 61.1 \\

FacTool
& 67.9 [61.9, 74.2]
& 67.9 [60.8, 74.5]
& 77.6 & 62.2 & 69.0 \\

Anchor-only LLM
& 71.9 [66.4, 77.4]
& 72.7 [65.9, 78.6]
& 83.6 & 59.0 & 69.2 \\

Source-agnostic LLM
& 75.9 [70.6, 81.2]
& 76.6 [69.8, 82.6]
& 88.3 & 62.8 & 73.4 \\

ModernBERT
& 76.2 [69.1, 82.6]
& 79.9 [71.6, 86.9]
& 61.2 & 96.2 & 74.8 \\

TF-IDF
& 77.9 [71.7, 83.3]
& 78.3 [69.9, 85.9]
& 56.9 & 100.0 & 72.6 \\

Cite-or-drop LLM
& 78.8 [73.5, 83.8]
& 77.6 [70.9, 83.6]
& 83.1 & 78.8 & 80.9 \\

SelfCheck-NLI
& 81.5 [75.2, 87.0]
& 85.1 [79.2, 90.3]
& 70.4 & 88.5 & 78.4 \\

DeBERTa-NLI
& 82.9 [77.4, 87.9]
& 89.4 [85.3, 92.9]
& 73.6 & 84.0 & 78.4 \\

\midrule

Representational
& 88.8 [84.6, 92.4]
& 90.8 [86.0, 94.5]
& 76.5 & 91.7 & 83.4 \\

Jacobian
& 89.4 [85.1, 93.4]
& 90.9 [86.9, 94.3]
& -- & -- & -- \\

AGR-value
& 89.8 [85.7, 93.3]
& 91.8 [87.2, 95.2]
& -- & -- & -- \\

\midrule

\textbf{AGR-slot}
& \textbf{96.2 [94.0, 97.9]}
& \textbf{97.3 [95.5, 98.6]}
& -- & -- & -- \\

\bottomrule
\end{tabular*}
\vspace{-7mm}
\end{table*}

\begin{wrapfigure}[12]{r}{0.40\linewidth}
\vspace{-.5cm}
\centering
\includegraphics[width=\linewidth]{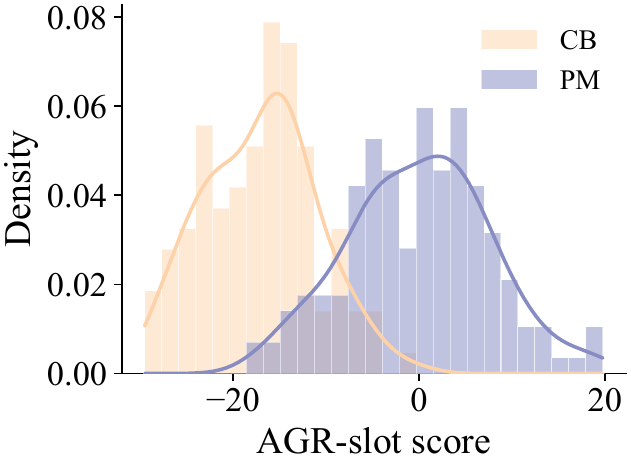}
\vspace{-.7cm}
\caption{AGR-slot score distribution on the Shopping test cohort.}
\label{fig:rq2-scoredist}
\end{wrapfigure}
\textbf{Baseline comparison.} Table~\ref{tab:rq2} compares AGR with lexical, embedding-based, and adapted factual verification methods on the same held-out Shopping cohort. AGR-slot achieves the highest detection performance, reaching 96.2\% AUROC and 97.3\% AUPRC. Compared with DeBERTa-NLI, the strongest external baseline in overall ranking performance, AGR-slot improves AUROC by 13.3 percentage points and AUPRC by 7.9 points. Its advantage extends across the lexical, embedding-based, and LLM verification baselines, including methods supplied with anchor-specific evidence. The representational readout alone also exceeds all evaluated external baselines in AUROC, indicating that the agent's pre-emission representations contain useful information for identifying PM. Incorporating functional information further improves discrimination,
suggesting that downstream computation provides predictive information
about the model's compatibility with attribute-slot targets beyond
what the linear representational probe captures on its own.

\textbf{Readout contributions.} Table~\ref{tab:rq2} further examines the contributions of the representational and Jacobian-based functional readouts. Individually, the two signals achieve comparable detection performance, with AUROC of 88.8\% and 89.4\%, respectively, demonstrating that both capture information relevant to PM prediction. Combining them yields additional improvements, most prominently in AGR-slot, which integrates the representational probe with slot-level functional information. Its AUROC increases to 96.2\%, accompanied by an AUPRC of 97.3\%. AGR-value, which incorporates anchor-value compatibility through calibrated fusion, also achieves competitive detection performance. The gain over both individual readouts suggests that similar
standalone accuracy can conceal complementary predictive information.

\textbf{Claim-level discrimination.} Figure~\ref{fig:rq2-scoredist} provides a complementary view of AGR-slot's detection performance through the score distributions of PM and correctly bound claims. The two groups exhibit distinct concentrations, with PM claims generally receiving higher scores, consistent with the strong ranking performance reported in Table~\ref{tab:rq2}. To further examine discrimination at comparable detection coverage, we evaluate the internal readouts at matched recall. At approximately 91.7\% recall, AGR-slot achieves 90.5\% precision, compared with 77.3\% for the representational readout and 77.7\% for AGR-value. This improvement corresponds to substantially fewer incorrectly flagged claims at the same PM detection coverage. On the 93 test claims with non-overlapping quote templates,
AGR-slot achieves 95.0\% AUROC, compared with 88.2\% for the
representational readout and 75.6\% for TF-IDF
(Appendix~\ref{app:template-overlap}).

\begin{wrapfigure}[11]{r}{0.48\linewidth}
\vspace{-.9cm}
\centering
\includegraphics[
    width=\linewidth
]{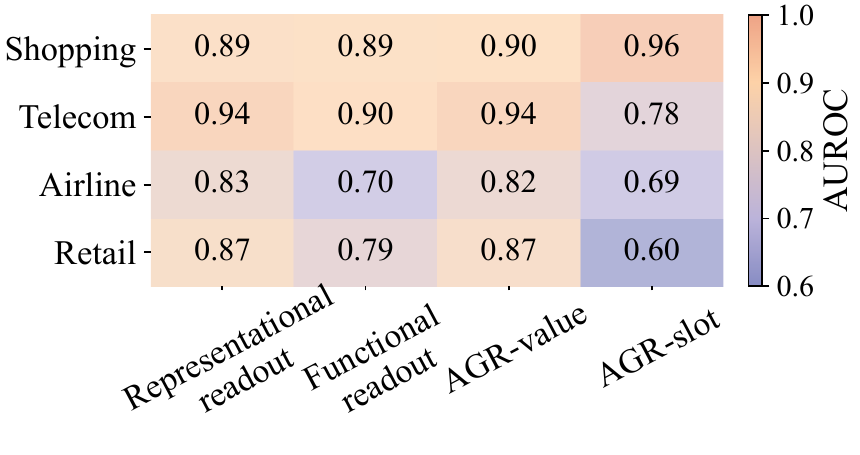}
\vspace{-0.9cm}
\caption{Cross-domain diagnostic AUROC of the evaluated readouts
under in-domain and zero-shot configurations.}
\label{fig:rq2-crossdomain}
\end{wrapfigure}
\textbf{Cross-domain analysis.} Figure~\ref{fig:rq2-crossdomain} extends the readout comparison to Telecom, Airline, and Retail. The representational readout achieves AUROC values of 0.94, 0.83, and 0.87 across these domains, while AGR-value achieves 0.94, 0.82, and 0.87, respectively. AGR-slot achieves its strongest performance on Shopping.
Across the three additional domains, the representational readout
outperforms the standalone functional readout, while AGR-value
closely matches the representational readout. This comparison
identifies the representational signal as the more consistently
discriminative component across these evaluated environments.

\subsection{Mitigation Effectiveness and Claim Preservation}
\label{sec:rq3}

We answer \textbf{RQ3} through end-to-end mitigation,
shared-gate correction, and detector-head ablation,
examining PM reduction and content preservation.

\begin{figure*}[t]
\centering
\includegraphics[
    width=\textwidth
]{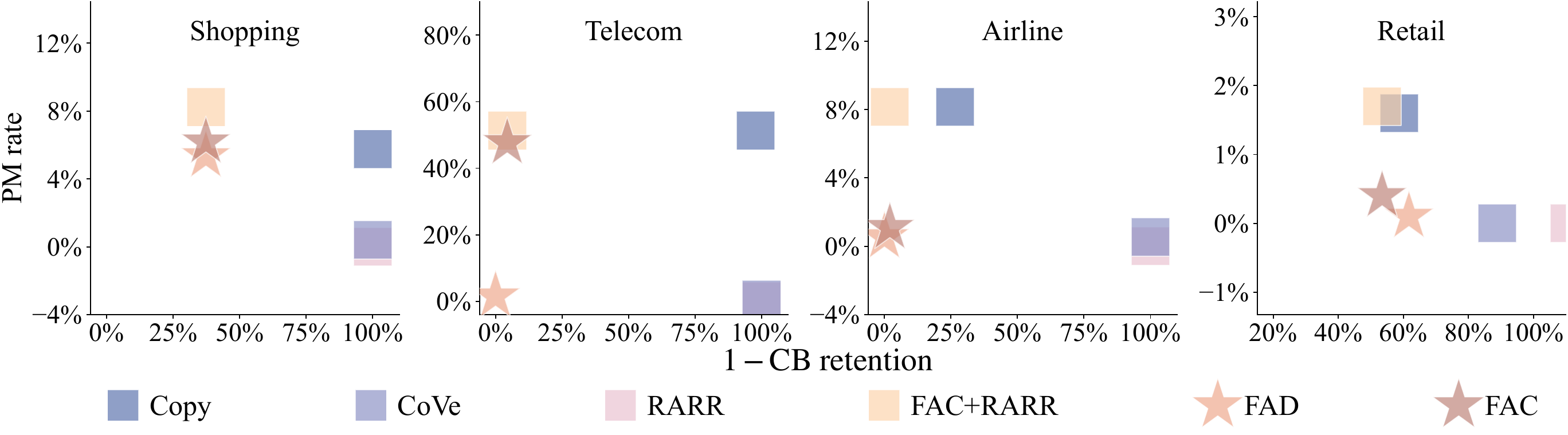}
\vspace{-0.7cm}
\caption{End-to-end mitigation across four agent domains.
Lower-left is better: FAC achieves a favorable trade-off
between PM reduction and correctly bound claim retention.}
\label{fig:mitigation-e2e}
\vspace{-0.2cm}
\end{figure*}
\textbf{End-to-end effectiveness.}
Figure~\ref{fig:mitigation-e2e} compares complete
intervention configurations across four domains.
On Shopping, FAC substantially reduces PM while retaining
correctly bound claims. Deletion-intensive alternatives
can achieve lower residual PM rates, but at the cost of
removing more valid content. FAC offers a different
trade-off by replacing unsupported attributes with
recoverable anchor-supported values rather than simply
discarding flagged claims.
The cross-domain results reveal an important distinction.
FAC also achieves substantial PM reduction on Airline
and Retail, whereas Telecom favors more aggressive
removal: Fixed-Anchor Deletion (FAD) achieves considerably
lower residual PM than the revision-based configuration.
This difference illustrates the limitations of revision
when suitable replacements cannot be recovered.
Consequently, mitigation performance depends not only
on identifying erroneous claims but also on the evidence
available for correction.

\begin{figure*}[t]
\centering
\includegraphics[
    width=\textwidth
]{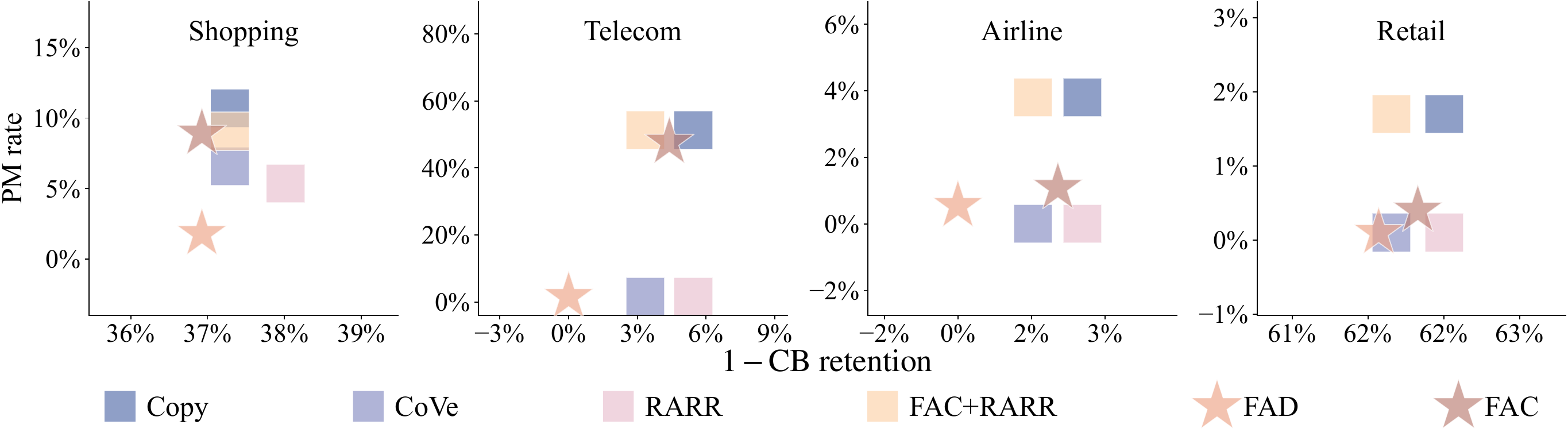}
\vspace{-.6cm}
\caption{Shared-gate mitigation across four agent domains. All methods receive identical claims flagged by a frozen representational probe, isolating differences in downstream routing and correction. Marker positions are jittered to reduce overlap. Exact residual PM rates are reported in Table~\ref{tab:shopping-mitigation-complete}}
\label{fig:mitigation-shared-gate}
\vspace{-.45cm}
\end{figure*}

\begin{wrapfigure}[12]{r}{0.48\linewidth}
\vspace{-0.7cm}
\centering
\includegraphics[
    width=\linewidth
]{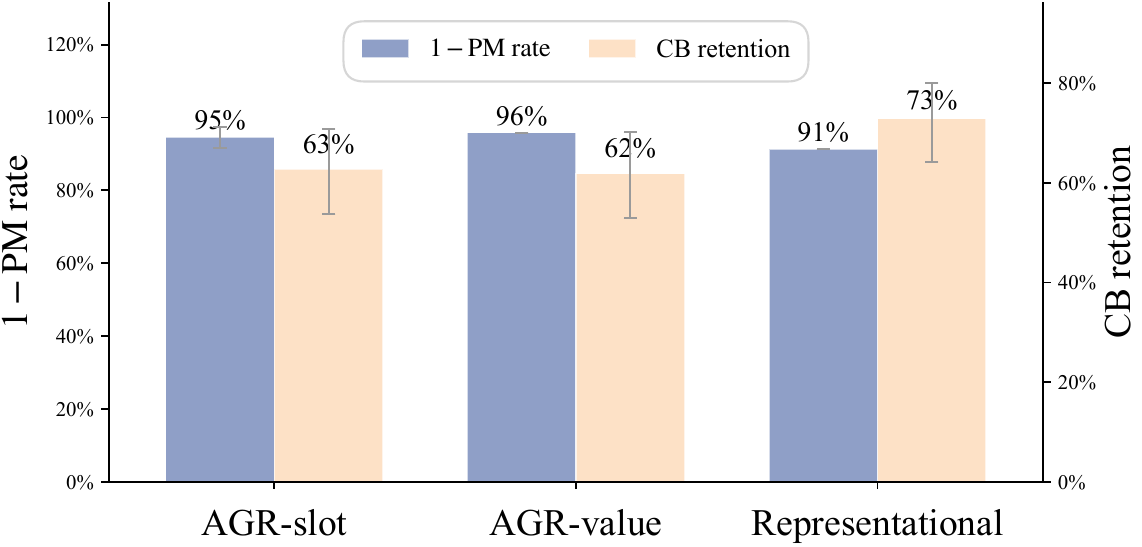}
\vspace{-0.7cm}
\caption{Detector-head comparison on Shopping with downstream
routing held fixed. Bars report the PM-free trajectory rate
and correctly bound claim retention at the operating points.}
\label{fig:detector-head-ablation}
\end{wrapfigure}
\textbf{Shared-gate correction.} To isolate downstream correction from claim selection, all methods
receive identical claims flagged by a frozen representational probe.
On Shopping, the shared gate is $p_{\mathrm{pm}}(c)>0.05$. Figure~\ref{fig:mitigation-shared-gate} compares their residual PM rates and correctly bound claim retention, while Appendix Table~\ref{tab:shopping-mitigation-complete} provides the complementary output-claim ratios. On Shopping, FAC, FAD, and RARR achieve the same residual PM rate of 5.3\% with comparable correctly bound claim retention. However, FAC preserves 59.3\% of the original output-claim volume, compared with 35.8\% for FAD and 37.0\% for RARR. This difference demonstrates the benefit of anchor-grounded revision over deletion-oriented alternatives at matched residual PM. Other methods exhibit different trade-offs: CoVe retains more output claims but leaves higher residual PM, while constrained copying retains substantially more content at the cost of additional binding errors. The cross-domain results further establish the importance of available anchor evidence. On Retail, FAC retains considerably more output content than FAD with a small increase in residual PM, whereas on Telecom, deletion achieves a much larger PM reduction than revision-based correction. Together, these controlled comparisons show that FAC can preserve additional response content without sacrificing measured PM reduction when supported replacements are available, while the removal branch remains important when anchor evidence is insufficient.

\paragraph{Detector-head ablation.}
Figure~\ref{fig:detector-head-ablation} compares upstream risk estimators with downstream routing
held fixed. At the evaluated operating points, AGR-value and
AGR-slot achieve PM-free trajectory rates of 96\% and 95\%,
with correctly bound claim retention of 62\% and 63\%, respectively.
The representational readout retains 73\% of correctly bound claims
at a PM-free trajectory rate of 91\%.
These results illustrate the trade-off between intervention
coverage and preservation of correctly bound claims.
This pattern is consistent with more aggressive intervention
under the AGR configurations, suppressing more PM while also
removing more correctly bound claims.

%% file: sections/06_related_work.tex
\section{Related Work}
\label{sec:related-work}

\vspace{-0.2cm}
\paragraph{Evaluation of LLM Agents.}
Recent agent benchmarks evaluate interactions with web environments~\citep{yao2022webshop,zhou2024webarena,deng2023mind2web}, external tools~\citep{li2023apibank,qin2024toolllm,lu2025toolsandbox,chen2025acebench}, and multi-step tasks~\citep{ma2024agentboard,trivedi2024appworld}. Domain-specific benchmarks further assess agent performance in realistic environments, including product recommendation~\citep{wang2025shoppingbench}. These evaluations primarily characterize task completion, interaction correctness, and tool-use capabilities. Related work on entity binding in tool-augmented agents investigates whether actions target the intended entities~\citep{babu_entity_2026,babu_binding_2026}. Our study examines a complementary failure that can occur after entity selection: whether claims in the final response remain grounded in observations associated with the committed entity.

\vspace{-0.2cm}
\paragraph{Factuality Verification and Correction.}
Existing factuality methods decompose generated responses into verifiable claims~\citep{akbar2024hallumeasure,hu2024refchecker,wei2024longform} and assess their consistency with external references, retrieved evidence, or generated verification questions~\citep{fabbri2022qafacteval,laban2022summac,honovich2022true,wang2024factcheckbench}. Attribution and citation-based approaches additionally connect generated claims to supporting evidence~\citep{gao2023alce,huang2024citation,zhang2025longcite}, while post-hoc correction methods revise unsupported content through evidence retrieval and verification~\citep{gao2023rarr}. Although these approaches provide mechanisms for assessing and improving factual support, applying them to multi-entity agent trajectories requires identifying the committed entity and enforcing its evidence boundary. We investigate this setting through claim-level anchor-specific verification and correction, explicitly preserving the original entity commitment.

\vspace{-0.2cm}
\paragraph{Entity Binding and Internal Representations.}
Studies of entity--attribute binding investigate how language models represent associations between entities and their properties~\citep{feng2024binding,dai2024binding}, while related research examines entity tracking and latent-state representations~\citep{kim2023entity,huang2025latentstate}. Representation-level analyses and model-editing methods further investigate the accessibility and localization of factual associations~\citep{meng2022rome,meng2023memit}. More recently, Jacobian-based approaches approximate downstream model computations to interpret information accessible from intermediate hidden states~\citep{gurnee_verbalizable_2026}. Building on these directions, we investigate entity-specific binding risk at the claim-generation state of an agent and combine representational and functional readouts for detection. Our setting connects internal representation analysis with evidence-grounded intervention over completed multi-entity trajectories.

%% file: sections/07_conclusion.tex
\section{Conclusion}
\label{sec:conclusion}
\vspace{-0.2cm}
We identify Phantom Merge (PM), an entity--evidence binding failure in which agents attribute unsupported claims to their committed entity. Evaluations across four domains and four backbones reveal diverse manifestations, including in task-correct trajectories. We introduce Anchor-Grounded Risk (AGR), combining representational and Jacobian-based functional readouts for claim-level detection, and Fixed-Anchor Correction (FAC), which revises or removes flagged claims using anchor evidence. Our experiments demonstrate effective PM detection and substantial reductions in binding errors while retaining response content. These findings establish entity-specific claim grounding as an important dimension of reliable multi-entity agents.

%% file: sections/appendix.tex
\clearpage
\section{Benchmark Construction and Annotation}
\label{app:annotation}
\counterwithin{table}{section}
\counterwithin{figure}{section}

\textbf{Evaluation cohorts.}
We evaluate ShoppingBench and the airline, telecom, and retail domains of $\tau^3$-Bench across four agent backbone families. The primary Shopping cohort contains 2,000 trajectories selected to emphasize multi-entity scenarios, while each $\tau^3$-Bench cohort contains 1,000 trajectories sampled from its task distribution. Shopping prevalence estimates therefore describe the constructed cohort. Detection uses a separately annotated Shopping dataset; its held-out test set contains 274 claims from 243 trajectories, with 156 PM and 118 correctly bound claims.

\textbf{Annotation protocol.}
The 2,000 Shopping trajectories were manually annotated according to Section~\ref{sec:problem}. Annotators identified the committed anchor, extracted anchor-attributed factual claims, and compared each claim with the user instruction, anchor evidence, and observations associated with other entities. They assigned the applicable \textsc{CEM}, \textsc{CAP}, \textsc{ASF}, and \textsc{PH} labels. All 274 held-out detection claims received an additional manual review. The Shopping data partition was established without reference to the observed PM label proportions. The detection test set evaluates PM against correctly bound claims; it does not include a separate PH negative class.

For the three $\tau^3$-Bench domains, OpenAI GPT-5.6 Astra performed two independent annotation passes. Each pass received the complete trajectory, user instruction, committed anchor, and evidence associated with all observed entities. Disagreements were reconciled, with human judgments taking precedence when needed. The same annotation criteria were used across domains and backbones.

\textbf{Human quality control.}
We audited 50 trajectories from each evaluated model--domain cohort of LLM-assisted annotations, stratifying the sample by PM-positive and PM-negative labels. Cohorts with unusually high PM prevalence received additional review. After reconciliation, no disagreements remained in the final PM labels of the inspected samples. Human review also clarified annotation wording and resolved disputed cases. This describes the final audited labels rather than a pre-adjudication agreement statistic.

\section{AGR Implementation and Readout Ablation}
\label{app:readout-ablation}

\textbf{Data partitions.}
The Shopping detection dataset contains 1,244 claims: 772 training, 198 development, and 274 test instances. For the AGR fitting protocol, the 970 non-test instances are partitioned into $\mathcal{D}_p$ ($n=393$) for probe fitting, $\mathcal{D}_c$ ($n=289$) for score calibration, and $\mathcal{D}_f$ ($n=288$) for fusion fitting and threshold selection. The shared held-out test set is unchanged. Parameters and selected thresholds remain fixed during test evaluation. The representational development split and $\mathcal{D}_f$ are identified separately below when reporting their respective operating points.

\textbf{Score definitions.}
Table~\ref{tab:detector-score-definitions} distinguishes the four detection outputs. The representational readout uses the frozen pre-emission hidden state. The standalone Jacobian score uses slot-level probability mass without the representational logit. AGR-slot combines that mass with the frozen probe logit, whereas AGR-value uses the fitted anchor-value fusion output described in Section~\ref{sec:method-agr}. All scores are oriented so that larger values indicate greater PM risk.

\begin{table}[H]
\centering
\small
\caption{Detection score identities. Here $\ell_{\mathrm{pm}}=\operatorname{logit}(p_{\mathrm{pm}})$, and $r_{\mathrm{val}}$ denotes the AGR-value fusion logit, including its fallback.}
\label{tab:detector-score-definitions}
\begin{tabular}{@{}p{0.20\linewidth}p{0.41\linewidth}p{0.32\linewidth}@{}}
\toprule
Readout & Evaluated score & Interpretation \\
\midrule
Representational & $p_{\mathrm{pm}}(c)$ & Frozen representational probe output \\
Jacobian (slot) & $\log\pi_J(T_{\mathrm{slot}}(c)\mid h_c)$ & Standalone slot-level score \\
AGR-value & $r_{\mathrm{val}}(c)$; $\rho(c)=\sigma(r_{\mathrm{val}}(c))$ & Fitted value-level fusion \\
AGR-slot & $B(c)=\ell_{\mathrm{pm}}(c)+\log\pi_J(T_{\mathrm{slot}}(c)\mid h_c)$ & Fixed-additive AGR-slot score \\
\bottomrule
\end{tabular}
\end{table}

\textbf{Anchor-value availability.}
The fitted AGR-value pipeline has an available anchor-value functional score for 23 instances in $\mathcal{D}_c$ and 48 of the 274 test claims. For the remaining 226 test claims, the fusion output falls back to the calibrated representational branch. All 274 claims contribute to the reported AGR-value metrics. These counts describe functional-score availability in the evaluated fusion pipeline.

\begin{wraptable}[9]{r}{0.37\linewidth}
\centering
\small
\caption{Readout comparison on Shopping.
AUROC is reported with its 95\% bootstrap confidence interval.}
\vspace{-0.2cm}
\label{tab:readout-ablation}
\setlength{\tabcolsep}{3pt}
\begin{tabular}{@{}lc@{}}
\toprule
Readout & AUROC [95\% CI] \\
\midrule
Representational & 0.888 [0.846, 0.924] \\
Jacobian (slot) & 0.894 [0.851, 0.934] \\
AGR-value & 0.898 [0.857, 0.933] \\
\textbf{AGR-slot} & \textbf{0.962 [0.940, 0.979]} \\
\bottomrule
\end{tabular}
\end{wraptable}
\textbf{Readout comparison.}
Table~\ref{tab:readout-ablation} compares the configurations on the same test cohort. Both individual readouts provide substantial discrimination. AGR-slot attains the highest AUROC, improving over the representational readout by 7.4 percentage points and over the standalone Jacobian score by 6.8 points. AGR-value provides a smaller improvement. The two fused configurations differ in both their functional target and fusion rule; these results establish their relative performance rather than isolate either design choice.

\begin{figure}[H]
\centering
\includegraphics[width=0.88\linewidth]{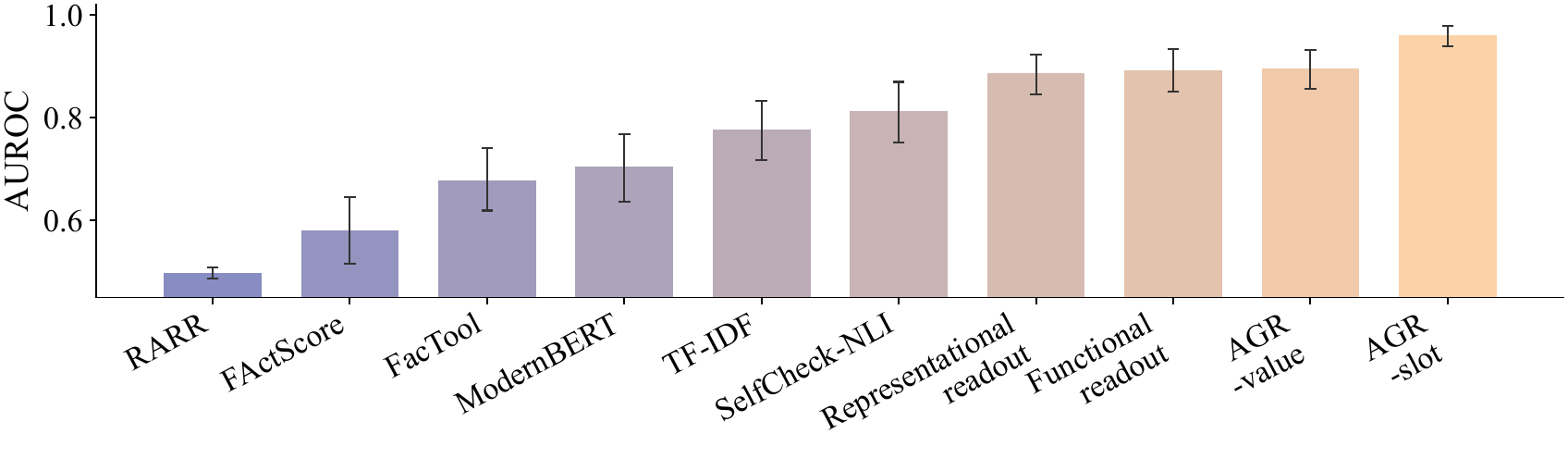}
\vspace{-.6cm}
\caption{AUROC of selected detection baselines and internal readouts. The ModernBERT variant shown in this plot is the quote-input variant.}
\label{fig:readout-ablation}
\end{figure}

\section{Detection Baselines and Complete Results}
\label{app:roc-overlay}

\textbf{Ranking performance.}
Table~\ref{tab:full-detection-registry} extends the main comparison with the additional ModernBERT and SelfCheckGPT configurations. All methods are evaluated on the same 274 test claims. AUROC and average precision (AP) summarize ranking performance. Confidence intervals for detection use 2,000 trajectory-level bootstrap replicates. The internal-score audit verified identical claim identifiers and labels across the representational and AGR test outputs.

The standalone Jacobian, AGR-value, and AGR-slot scores reproduce AUROC/AP values of 0.8942/0.9089, 0.8976/0.9179, and 0.9618/0.9730, respectively. AGR-slot's result uses the AGR-slot score in Table~\ref{tab:detector-score-definitions}.

\begin{table}[H]
\centering
\caption{Extended detection results on the Shopping test cohort ($n=274$). All metrics are on a 0--1 scale; AUROC and AP include 95\% bootstrap confidence intervals. P/R/F1 use the common threshold of 0.05 for the designated probability-based methods. ``-- denotes internal scores whose operating points are reported in Appendix~\ref{app:threshold-selection}. SC denotes SelfCheckGPT-NLI and MB denotes ModernBERT. $\dagger$ marks predictions that label every test claim as PM.}
\label{tab:full-detection-registry}
\small
\setlength{\tabcolsep}{3pt}
\renewcommand{\arraystretch}{1.08}
\begin{tabular*}{\linewidth}{@{\extracolsep{\fill}}lccrrr@{}}
\toprule
Method & AUROC [95\% CI] & AP [95\% CI] & P & R & F1 \\
\midrule
RARR gate
& 0.499 [0.487, 0.508]
& 0.569 [0.502, 0.630]
& 0.500 & 0.006 & 0.013 \\
FActScore
& 0.582 [0.516, 0.646]
& 0.655 [0.575, 0.730]
& 0.594 & 0.628 & 0.611 \\
FacTool
& 0.679 [0.619, 0.742]
& 0.679 [0.608, 0.745]
& 0.776 & 0.622 & 0.690 \\
SC-NLI ($K=5$)
& 0.693 [0.626, 0.757]
& 0.752 [0.676, 0.817]
& 0.598 & 0.974 & 0.741 \\
MB (quote)
& 0.706 [0.637, 0.769]
& 0.729 [0.636, 0.809]
& 0.617 & 0.962 & 0.752 \\
MB (structured)
& 0.762 [0.691, 0.826]
& 0.799 [0.716, 0.869]
& 0.612 & 0.962 & 0.748 \\
TF-IDF
& 0.779 [0.717, 0.833]
& 0.783 [0.699, 0.859]
& 0.569 & 1.000$^\dagger$ & 0.726 \\
SC-NLI ($K=1$)
& 0.815 [0.752, 0.870]
& 0.851 [0.792, 0.903]
& 0.704 & 0.885 & 0.784 \\
Anchor-only LLM
& 0.719 [0.664, 0.774] & 0.727 [0.659, 0.786] & 0.836 & 0.590 & 0.692 \\
Source-agnostic LLM
& 0.759 [0.706, 0.812] & 0.766 [0.698, 0.826] & 0.883 & 0.628 & 0.734 \\
Cite-or-drop LLM
& 0.788 [0.735, 0.838] & 0.776 [0.709, 0.836] & 0.831 & 0.788 & 0.809 \\
DeBERTa-MNLI
& 0.829 [0.774, 0.879] & 0.894 [0.853, 0.929] & 0.736 & 0.840 & 0.784 \\
\midrule
Representational
& 0.888 [0.846, 0.924]
& 0.908 [0.860, 0.945]
& 0.765 & 0.917 & 0.834 \\
Jacobian (slot)
& 0.894 [0.851, 0.934]
& 0.909 [0.869, 0.943]
& -- & -- & -- \\
AGR-value
& 0.898 [0.857, 0.933]
& 0.918 [0.872, 0.952]
& -- & -- & -- \\
\midrule
\textbf{AGR-slot}
& \textbf{0.962 [0.940, 0.979]}
& \textbf{0.973 [0.955, 0.986]}
& -- & -- & -- \\
\bottomrule
\end{tabular*}
\end{table}

\begin{wrapfigure}{r}{0.40\linewidth}
\vspace{-0.4cm}
\centering
\includegraphics[width=\linewidth]{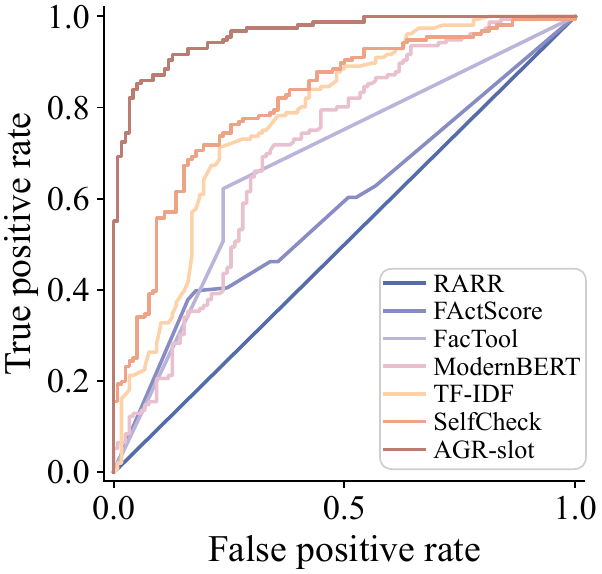}
\caption{ROC curves for selected detection methods on the Shopping test set.}
\label{fig:roc-overlay}
\end{wrapfigure}
\textbf{Recall-matched comparison.}
At approximately 91.7\% recall, AGR-slot achieves 90.5\% precision, compared with 77.3\% for the representational readout, 77.7\% for AGR-value, and 64.7\% for the standalone Jacobian score. These are descriptive operating points obtained by matching recall on the test predictions. The representational precision differs from its fixed-threshold value because recall matching selects a different operating point. Development-selected thresholds are evaluated separately in Appendix~\ref{app:threshold-selection}.

\textbf{Paired ranking comparison.}
A paired DeLong comparison~\citep{delong_comparing_1988} between AGR-slot and AGR-value yields $p=4.83\times10^{-16}$. This comparison pairs detector scores by claim. The trajectory-level bootstrap intervals above account for grouping by trajectory; the standard DeLong test does not additionally model within-trajectory dependence.

\subsection{Detection by Quote-Template Overlap}
\label{app:template-overlap}

We stratify the test cohort using the quote-fingerprint protocol underlying the reported 66.1\% training-template overlap. Of the 274 test claims, 181 share a quote template with the training set and 93 do not. The detectors remain frozen, and TF-IDF is fitted on training data. No retraining or threshold adjustment is performed for this analysis. Confidence intervals use 2,000 trajectory-level bootstrap replicates.

\begin{table}[H]
\centering
\small
\setlength{\tabcolsep}{3pt}
\caption{Detection stratified by training-set quote-template overlap. Cells report AUROC [95\% CI]; PM+ is the number of PM-positive claims.}
\label{tab:template-stratified}
\begin{tabular*}{\linewidth}{@{\extracolsep{\fill}}lrrccc@{}}
\toprule
Subset & $n$ & PM+ & AGR-slot & Representational & TF-IDF \\
\midrule
Overlap & 181 & 108 & .972 [.950, .989] & .903 [.860, .942] & .783 [.706, .851] \\
Non-overlap & 93 & 48 & .950 [.907, .984] & .882 [.805, .941] & .756 [.645, .855] \\
\bottomrule
\end{tabular*}
\end{table}

AGR-slot maintains an AUROC of 0.950 on the non-overlapping subset, compared with 0.882 for the representational readout and 0.756 for TF-IDF. Its advantage is therefore not confined to claims sharing a quote template with training examples. This exploratory analysis measures overlap under the specified fingerprint protocol, rather than all lexical or semantic similarities.

\section{Detection Threshold Selection}
\label{app:threshold-selection}
\label{app:operating-points}

\begin{wrapfigure}[13]{r}{0.40\linewidth}
\vspace{-0.9cm}
\centering
\includegraphics[width=1\linewidth]{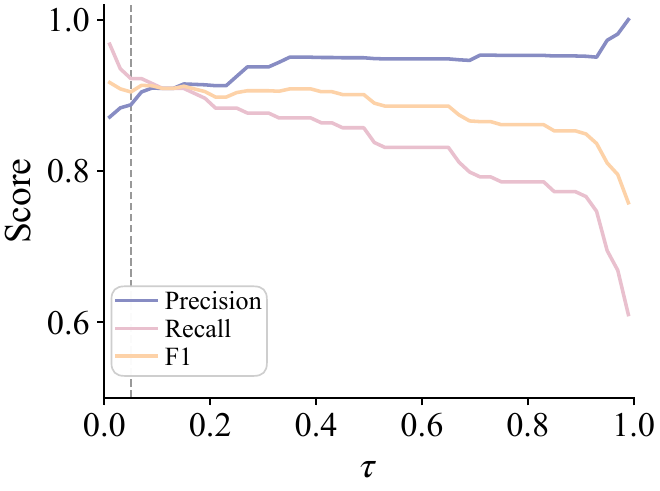}
\vspace{-1cm}
\caption{Representational-gate precision, recall, and F1 on development set. The marked threshold of 0.05 is used for Shopping shared-gate claim selection.}
\label{fig:threshold-sweep}
\end{wrapfigure}

\textbf{Common-threshold protocol.}
The main P/R/F1 comparison uses a common threshold of 0.05 for the designated probability-based methods. The Shopping shared-gate correction experiments use the frozen representational probe at this threshold. Figure~\ref{fig:threshold-sweep} shows its development-set sensitivity. At 0.05, TF-IDF predicts all 274 test claims as PM, yielding recall of 1.000. The internal ranking scores are evaluated below at their own development-selected operating points.

\textbf{Development-selected operating points.}
Table~\ref{tab:internal-operating-points} reports thresholds selected without test labels. The Jacobian threshold maximizes F1 on the representational development split. The $\mathcal{D}_f$ thresholds maximize F1 on the AGR fusion split ($n=288$). For AGR-value, this split also fits the fusion weights, so it is not an independent threshold-validation set. We report both saved AGR-slot thresholds, identifying their respective selection splits. These results supplement the common-threshold comparison; external baselines have not been uniformly retuned under this protocol.

The Jacobian operating point detects 120 PM claims with no false positives among the 118 correctly bound claims. Its 100\% precision is the observed value on this finite test cohort. The two AGR-slot thresholds expose a precision--recall trade-off: the $\mathcal{D}_f$ threshold yields higher precision, while the representational development threshold yields higher recall, with similar F1 scores.

\textbf{AGR-value on the probability scale.}
Applying $\rho(c)=\sigma(r_{\mathrm{val}}(c))\geq0.05$ yields 61.3\% precision, 99.4\% recall, and 75.8\% F1, with $(TP,FP,FN,TN)=(155,98,1,20)$. This is a computable probability-scale operating point and strongly favors recall. The dashes for AGR-value in the main table denote separate operating-point reporting, rather than an inability to compute P/R/F1.

\begin{table}[H]
\centering
\small
\caption{Internal detection operating points on the full Shopping test cohort. P/R/F1 are percentages; thresholds use the raw score scales in Table~\ref{tab:detector-score-definitions}.}
\label{tab:internal-operating-points}
\begin{tabular}{@{}llrrrr@{}}
\toprule
Method & Selection split & Threshold & P & R & F1 \\
\midrule
Jacobian (slot) & representational dev & $-7.064$ & 100.0 & 76.9 & 87.0 \\
AGR-value & $\mathcal{D}_f$ & $-0.7585$ & 76.4 & 93.6 & 84.1 \\
AGR-slot & $\mathcal{D}_f$ & $-8.822$ & 91.9 & 87.2 & 89.5 \\
AGR-slot & representational dev & $-12.714$ & 84.5 & 94.2 & 89.1 \\
\bottomrule
\end{tabular}
\end{table}

\begin{table}[H]
\centering
\small
\caption{Confusion counts for the operating points in Table~\ref{tab:internal-operating-points}. PM is the positive class.}
\label{tab:internal-confusion-counts}
\begin{tabular}{@{}llrrrr@{}}
\toprule
Method & Selection split & TP & FP & FN & TN \\
\midrule
Jacobian (slot) & representational dev & 120 & 0 & 36 & 118 \\
AGR-value & $\mathcal{D}_f$ & 146 & 45 & 10 & 73 \\
AGR-slot & $\mathcal{D}_f$ & 136 & 12 & 20 & 106 \\
AGR-slot & representational dev & 147 & 27 & 9 & 91 \\
\bottomrule
\end{tabular}
\end{table}

\section{Baseline Implementation and Adaptation}
\label{app:baseline-details}

Table~\ref{tab:baserepresentationaldaptation} summarizes the implementations and experimental roles of the baselines. Detection methods assess claim-level risk; mitigation methods additionally implement a correction procedure.

\begin{table}[H]
\centering
\caption{Implementation provenance and experimental roles of the evaluated baselines.}
\label{tab:baserepresentationaldaptation}
\small
\setlength{\tabcolsep}{3pt}
\renewcommand{\arraystretch}{1.1}
\begin{tabular}{@{}p{0.19\linewidth}p{0.75\linewidth}@{}}
\toprule
Method & Implementation and role \\
\midrule
RARR
& Adapted from pinned third-party prompts and executed against trajectory observations using our generation backend. Its agreement gate is evaluated for detection; its revision procedure is evaluated for mitigation. \\
\addlinespace
CoVe
& CoVe-inspired verification and revision procedure implemented from the published method description. Evaluated as a mitigation baseline. \\
\addlinespace
Constrained copy
& Self-designed, FEVER-style constrained-copy baseline. Evaluated for mitigation under the corresponding intervention protocols. \\
\addlinespace
SelfCheckGPT, FacTool, FActScore
& Adapted detection implementations using the available trajectory observations. Evaluated as detection baselines, without a corresponding mitigation implementation. \\
\addlinespace
ModernBERT
& Embedding-based classifier trained on the labeled split. Quote-based and structured-input variants are reported. \\
\addlinespace
DeBERTa-MNLI
& Anchor-aware verification implementation based on an NLI checkpoint. Evaluated for detection and verification, without a separate rewriting procedure. \\
\addlinespace
LLM verifiers
& Anchor-only, source-agnostic, and cite-or-drop verification configurations. Evaluated for claim-level detection and verification, without a rewriting backend. \\
\bottomrule
\end{tabular}
\end{table}

\textbf{Intervention protocols.}
The Shopping complete-intervention configurations of RARR and CoVe perform an exhaustive intervention scan using the adapted backends. In shared-gate experiments, every correction backend receives the same externally selected claims. This separates complete-system performance from correction behavior conditional on a common claim selection.

\textbf{FAC implementation.}
Shopping shared-gate experiments use the frozen representational probe with a development-selected threshold of $\theta=0.05$. Claims satisfying $p_{\mathrm{pm}}(c)>\theta$ enter FAC. This gate is the representational probe, not the AGR-slot AGR-slot score. FAC uses precomputed audit states derived from the frozen probe, slot adjudication, and Jacobian-lens readouts with top-$k=50$. Functional audit thresholds are fitted on development data. In the end-to-end configuration, audit states are constructed during inference and Qwen3-8B performs evidence-grounded rewriting. The committed anchor is fixed throughout correction.

\begin{table}[H]
\centering
\small
\caption{Score and threshold roles in Shopping detection and correction. Detection operating points and audit thresholds have different roles.}
\label{tab:score-gate-roles}
\begin{tabular}{@{}p{0.28\linewidth}p{0.29\linewidth}p{0.4\linewidth}@{}}
\toprule
Role & Score or configuration & Rule \\
\midrule
Primary detection ranking & AGR-slot $B(c)$ & Threshold-independent AUROC/AP \\
Shared-gate claim selection & Representational $p_{\mathrm{pm}}(c)$ & Flag if $p_{\mathrm{pm}}(c)>0.05$ \\
Standalone Jacobian detection & Slot log mass & representational dev F1 threshold $-7.064$ \\
Jacobian audit partition & Slot log mass & Development PM-positive median $-3.401$ \\
Functional audit readout & Slot lexicon, top-$k=50$ & Development-fitted audit thresholds \\
Evidence-grounded rewriting & Qwen3-8B & Deterministic decoding; fixed anchor \\
\bottomrule
\end{tabular}
\end{table}

\section{Complete Mitigation Results}
\label{app:crossdomain-mitigation}

Tables~\ref{tab:e2e-crossdomain} and~\ref{tab:sharedgate-crossdomain} report complete-intervention and shared-gate results. Residual PM is measured at the trajectory level on the corresponding mitigation cohorts, whose initial rates differ from the characterization cohorts in Section~\ref{sec:rq1}. Mitigation confidence intervals use 500 bootstrap replicates.

Each cell reports residual PM with its 95\% confidence interval and the output-to-original claim ratio $R$. This ratio measures output claim count, not retention of the original correctly bound claims, and can exceed one when an intervention generates additional claims. Zero-event bootstrap intervals can collapse to $[0,0]$; they do not establish zero population risk.

\providecommand{\PMcell}[4]{}
\renewcommand{\PMcell}[4]{%
  \ensuremath{#1\%\,[#2,#3],\;R=#4}%
}
\begin{table}[H]
\centering
\caption{Complete-intervention mitigation across the three $\tau^3$-Bench domains. Initial PM rates are 52.4\% for Telecom, 20.1\% for Airline, and 3.1\% for Retail. Each cell reports residual PM [95\% CI] and the output-claim ratio $R$. FAC denotes Fixed-Anchor Correction and FAD denotes Fixed-Anchor Deletion.}
\label{tab:e2e-crossdomain}
\small
\setlength{\tabcolsep}{4pt}
\renewcommand{\arraystretch}{1.12}
\begin{tabular*}{\linewidth}{@{\extracolsep{\fill}}lccc@{}}
\toprule
Method & Telecom & Airline & Retail \\
\midrule
FAD
& \PMcell{1.6}{0.0}{3.7}{0.49}
& \PMcell{0.5}{0.0}{1.6}{0.81}
& \PMcell{0.1}{0.0}{0.3}{0.37} \\
RARR
& \PMcell{0.0}{0.0}{0.0}{0.00}
& \PMcell{0.0}{0.0}{0.0}{0.00}
& \PMcell{0.0}{0.0}{0.0}{0.00} \\
CoVe
& \PMcell{0.5}{0.0}{2.1}{0.01}
& \PMcell{0.5}{0.0}{1.6}{0.01}
& \PMcell{0.0}{0.0}{0.0}{0.00} \\
Constrained copy
& \PMcell{51.3}{44.5}{58.4}{1.30}
& \PMcell{8.2}{4.3}{12.0}{1.01}
& \PMcell{1.6}{0.9}{2.4}{0.91} \\
FAC
& \PMcell{47.6}{40.3}{55.0}{1.01}
& \PMcell{1.1}{0.0}{2.7}{0.79}
& \PMcell{0.4}{0.1}{0.8}{0.75} \\
FAC + RARR
& \PMcell{51.3}{44.5}{58.4}{1.30}
& \PMcell{8.2}{4.3}{12.0}{1.01}
& \PMcell{1.7}{1.0}{2.5}{0.78} \\
\bottomrule
\end{tabular*}
\end{table}

\begin{table}[H]
\centering
\caption{Shared-gate mitigation across the same three domains. All methods receive identical flagged claims within each domain. Each cell reports residual PM [95\% CI] and the output-claim ratio $R$.}
\label{tab:sharedgate-crossdomain}
\small
\setlength{\tabcolsep}{4pt}
\renewcommand{\arraystretch}{1.12}
\begin{tabular*}{\linewidth}{@{\extracolsep{\fill}}lccc@{}}
\toprule
Method & Telecom & Airline & Retail \\
\midrule
FAD
& \PMcell{1.6}{0.0}{3.7}{0.49}
& \PMcell{0.5}{0.0}{1.6}{0.81}
& \PMcell{0.1}{0.0}{0.3}{0.37} \\
RARR
& \PMcell{1.6}{0.0}{3.7}{0.47}
& \PMcell{0.0}{0.0}{0.0}{0.78}
& \PMcell{0.1}{0.0}{0.3}{0.37} \\
CoVe
& \PMcell{1.6}{0.0}{3.7}{0.47}
& \PMcell{0.0}{0.0}{0.0}{0.78}
& \PMcell{0.1}{0.0}{0.3}{0.37} \\
Constrained copy
& \PMcell{51.3}{44.5}{58.4}{1.30}
& \PMcell{3.8}{1.6}{6.5}{1.01}
& \PMcell{1.7}{1.0}{2.5}{0.91} \\
FAC
& \PMcell{47.6}{40.3}{55.0}{1.01}
& \PMcell{1.1}{0.0}{2.7}{0.79}
& \PMcell{0.4}{0.1}{0.8}{0.75} \\
FAC + RARR
& \PMcell{51.3}{44.5}{58.4}{1.30}
& \PMcell{3.8}{1.6}{6.5}{1.01}
& \PMcell{1.7}{1.0}{2.5}{0.78} \\
Audit-only
& \PMcell{49.7}{42.9}{57.1}{0.83}
& \PMcell{3.8}{1.6}{6.5}{0.22}
& \PMcell{1.6}{0.9}{2.4}{0.54} \\
\bottomrule
\end{tabular*}
\end{table}

FAC reduces residual PM from 20.1\% to 1.1\% on Airline and from 3.1\% to 0.4\% on Retail, with output-claim ratios of 0.79 and 0.75, respectively. On Telecom, FAC leaves 47.6\% residual PM, compared with 1.6\% for FAD. The relative value of revision and deletion therefore varies substantially by domain. The complete-intervention RARR configuration produces an output-claim ratio of 0.00 at the reported precision across these domains, so its zero residual PM must be read alongside the corresponding content loss.

\section{Rewrite-Backend Ablation}
\label{app:rewrite-backend-ablation}

We replace FAC's rewriting procedure with the adapted RARR backend while retaining the intervention configuration. Table~\ref{tab:shopping-mitigation-complete} reports this comparison alongside the other Shopping mitigation methods. Complete-intervention and shared-gate results are shown separately because they use different detection configurations.

\begin{table}[H]
\centering
\caption{Complete Shopping mitigation results. PM is residual trajectory-level PM rate with 95\% confidence interval. $R$ denotes the output-claim ratio. The initial PM rate is 51.4\%.}
\label{tab:shopping-mitigation-complete}
\small
\setlength{\tabcolsep}{4pt}
\renewcommand{\arraystretch}{1.08}
\begin{tabular*}{\linewidth}{@{\extracolsep{\fill}}lcccc@{}}
\toprule
& \multicolumn{2}{c}{Complete intervention}
& \multicolumn{2}{c}{Shared gate} \\
\cmidrule(lr){2-3}
\cmidrule(l){4-5}
Method & PM [95\% CI] & $R$ & PM [95\% CI] & $R$ \\
\midrule
FAD
& 5.3 [2.5, 8.2] & 0.358
& 5.3 [2.5, 8.2] & 0.358 \\
RARR
& 0.0 [0.0, 0.0] & 0.041
& 5.3 [2.5, 8.2] & 0.370 \\
CoVe
& 0.4 [0.0, 1.2] & 0.572
& 6.6 [3.7, 9.7] & 0.691 \\
Constrained copy
& 5.8 [2.9, 9.1] & 1.128
& 10.7 [6.6, 14.6] & 0.947 \\
FAC
& 6.2 [3.3, 9.1] & 0.807
& 5.3 [2.5, 8.2] & 0.593 \\
FAC + RARR
& 8.2 [4.5, 11.5] & 0.926
& 9.1 [5.3, 12.3] & 0.926 \\
\bottomrule
\end{tabular*}
\end{table}

Under complete intervention, FAC reduces trajectory-level PM from 51.4\% to 6.2\% with an output-claim ratio of 0.807. Replacing its rewrite backend with RARR yields a higher output-claim ratio of 0.926 and a higher residual PM rate of 8.2\%. The shared-gate comparison shows the same direction of change: FAC has 5.3\% residual PM and $R=0.593$, compared with 9.1\% and $R=0.926$ for FAC+RARR.

\begin{wraptable}[9]{r}{0.35\linewidth}
\vspace{-.8cm}
\centering
\small
\caption{Shopping shared-gate action counts:
187 flagged and 87 unflagged claims.}
\vspace{-.3cm}
\label{tab:fac-branch-accounting}
\setlength{\tabcolsep}{3pt}
\begin{tabular}{@{}lr@{}}
\toprule
Category & Claims \\
\midrule
Unflagged / keep & 87 \\
Flagged: copy-rewrite & 57 \\
Flagged: probe-only deletion & 125 \\
Flagged: remaining cases & 5 \\
\midrule
Total & 274 \\
\bottomrule
\end{tabular}
\end{wraptable}
The shared-gate results also separate revision from deletion-oriented correction. FAC, FAD, and RARR all have 5.3\% residual PM, while their output-claim ratios are 0.593, 0.358, and 0.370. FAC thus produces more output claims at the same measured residual PM rate. Correctly bound claim retention is evaluated separately in the main mitigation figures.

\textbf{Shopping branch accounting.}
At the shared probe threshold of 0.05, 187 of the 274 test claims are flagged and 87 are unflagged. The action summary records 57 flagged claims in the copy-rewrite branch and 125 in the probe-only deletion branch. Table~\ref{tab:fac-branch-accounting} reports these counts and the five remaining flagged cases separately. All 57 rewritten claims were manually verified to be supported by anchor evidence.

\section{Anchor-Aware Verification}
\label{app:anchor-verification}

Table~\ref{tab:anchor-verification} evaluates anchor-aware verification on the held-out Shopping test cohort. PM recall measures the fraction of PM claims flagged. CB drop measures the fraction of correctly bound claims flagged for removal, and CB retention is its complement. The frozen representational probe provides the common-threshold reference for shared-gate claim selection.

\begin{table}[H]
\centering
\caption{Anchor-aware verification on the held-out Shopping test cohort ($n=274$). PM R denotes PM recall; CB drop and CB ret. denote correctly bound claim false-drop rate and retention, respectively.}
\label{tab:anchor-verification}
\small
\setlength{\tabcolsep}{3pt}
\renewcommand{\arraystretch}{1.1}
\begin{tabular*}{\columnwidth}{@{\extracolsep{\fill}}lccc@{}}
\toprule
Verifier & PM R & CB drop & CB ret. \\
\midrule
Anchor-only LLM     & 0.59 & 0.15 & 0.85 \\
Source-agnostic LLM & 0.63 & 0.11 & 0.89 \\
Cite-or-drop LLM    & 0.79 & 0.21 & 0.79 \\
DeBERTa-MNLI        & 0.84 & 0.40 & 0.60 \\
Probe gate          & 0.92 & 0.37 & 0.63 \\
\midrule
Lexical anchor-only     & 0.71 & 0.40 & 0.60 \\
Lexical source-agnostic & 0.58 & 0.37 & 0.63 \\
Lexical cite-or-drop    & 0.63 & 0.39 & 0.61 \\
\bottomrule
\end{tabular*}
\end{table}

Cite-or-drop LLM attains 79\% PM recall with 79\% correctly bound claim retention, while the source-agnostic LLM retains 89\% of correctly bound claims at 63\% PM recall. DeBERTa-MNLI reaches 84\% recall with 60\% retention. The probe gate reaches 92\% recall with 63\% retention. These results characterize verification trade-offs before applying FAC's downstream audit and correction. The lexical variants provide additional controls under the corresponding verification configurations.

\section{Limitations}
\label{sec:limitations}

Our analysis concerns completed trajectories with recoverable entity commitments and observed entity-specific evidence. FAC preserves the committed anchor and therefore does not correct erroneous initial entity selection or earlier tool actions. Its ability to revise a claim depends on recovering supporting anchor evidence. The internal readouts provide predictive evidence about PM risk rather than establish a causal mechanism for the original failure.